\documentclass{article}

\usepackage{arxiv}

\usepackage[utf8]{inputenc} 
\usepackage[T1]{fontenc}    
\usepackage{hyperref}       
\usepackage{url}            
\usepackage{booktabs}       
\usepackage{amsfonts}       
\usepackage{amssymb}        
\usepackage{amsmath}        
\usepackage{nicefrac}       
\usepackage{microtype}      
\usepackage{cleveref}       
\usepackage{graphicx}
\usepackage{natbib}
\usepackage{doi}
\usepackage{algorithm}
\usepackage{algorithmic}

\title{Review of Case-Based Reasoning for LLM Agents: Theoretical Foundations, Architectural Components, and Cognitive Integration}

\newif\ifuniqueAffiliation
\uniqueAffiliationtrue

\ifuniqueAffiliation 
\author{ \href{https://orcid.org/0000-0001-8191-5728}{\includegraphics[scale=0.06]{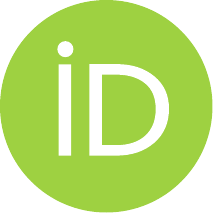}\hspace{1mm}Kostas Hatalis}\thanks{Corresponding author: kostas@gocharlie.ai} \\
	GoCharlie.ai\\
	\texttt{kostas@gocharlie.ai} \\
	\And
	\href{https://orcid.org/0000-0001-9359-8926}{\includegraphics[scale=0.06]{orcid.pdf}\hspace{1mm}Despina Christou} \\
	GoCharlie.ai\\
	\texttt{despina@gocharlie.ai} \\
    \And
	\href{https://orcid.org/0009-0003-3489-2895}{\includegraphics[scale=0.06]{orcid.pdf}\hspace{1mm}Vyshnavi Kondapalli} \\
	GoCharlie.ai\\
	\texttt{vyk223@lehigh.edu} \\
}
\else
\usepackage{authblk}

\newbox{\orcid}\sbox{\orcid}{\includegraphics[scale=0.06]{orcid.pdf}} 
\author[1]{%
	\href{https://orcid.org/0000-0000-0000-0000}{\usebox{\orcid}\hspace{1mm}Kostas Hatalis\thanks{\texttt{kostas@gocharlie.ai}}}%
}
\author[1]{%
	\href{https://orcid.org/0000-0000-0000-0000}{\usebox{\orcid}\hspace{1mm}Despina Christou\thanks{\texttt{despina@gocharlie.ai}}}%
}
\affil[1]{GoCharlie.ai}
\fi

\renewcommand{\shorttitle}{CBR for LLM Agents}

\hypersetup{
pdftitle={Case-Based Reasoning for Large Language Model Agents: Theoretical Foundations, Architectural Components, and Cognitive Integration},
pdfsubject={cs.AI, cs.CL, cs.LG},
pdfauthor={Kostas Hatalis, Despina Christou},
pdfkeywords={Case-Based Reasoning \and Large Language Models \and Retrieval-Augmented Generation \and Chain-of-Thought Reasoning \and Autonomous Agents \and Goal-Driven Autonomy \and Cognitive Architecture},
}

\begin{document}
\maketitle
 
\begin{abstract}
    Agents powered by Large Language Models (LLMs) have recently demonstrated impressive capabilities in various tasks. Still, they face limitations in tasks requiring specific, structured knowledge, flexibility, or accountable decision-making. While agents are capable of perceiving their environments, forming inferences, planning, and executing actions towards goals, they often face issues such as hallucinations and lack of contextual memory across interactions. This paper explores how Case-Based Reasoning (CBR), a strategy that solves new problems by referencing past experiences, can be integrated into LLM agent frameworks. This integration allows LLMs to leverage explicit knowledge, enhancing their effectiveness. We systematically review the theoretical foundations of these enhanced agents, identify critical framework components, and formulate a mathematical model for the CBR processes of case retrieval, adaptation, and learning. We also evaluate CBR-enhanced agents against other methods like Chain-of-Thought reasoning and standard Retrieval-Augmented Generation, analyzing their relative strengths. Moreover, we explore how leveraging CBR's cognitive dimensions (including self-reflection, introspection, and curiosity) via goal-driven autonomy mechanisms can further enhance the LLM agent capabilities. Contributing to the ongoing research on neuro-symbolic hybrid systems, this work posits CBR as a viable technique for enhancing the reasoning skills and cognitive aspects of autonomous LLM agents.
\end{abstract}


\section{Introduction}
\label{sec:intro}

Groundbreaking advances in intelligence have been driven by LLMs showcasing remarkable capabilities in comprehending language and carrying out tasks effectively. The development of systems empowered by LLM technology marks a frontier in AI research with significant impacts on the collaboration between humans and AI. Despite their abilities and skills, these agents face difficulties in intricate reasoning situations that require specialized knowledge in the field, and they need to improve in how they apply that knowledge and explain the reasoning behind their decisions. \citep{huang2023towards}. 

CBR, a concept rooted in both computer science and AI, presents a method to tackle these challenges. CBR operates on the idea that similar problems tend to have similar solutions by using experiences and knowledge stored in previous cases to solve new problems \citep{aamodt1994case}. This approach closely resembles how human experts often rely on reasoning and past decisions to solve problems \citep{kolodner1992introduction}. The standard CBR process includes four main steps: (1) Retrieve relevant cases; (2) Reuse the knowledge embedded in these cases; (3) Revise the proposed solution; and (4) Retain the new problem-solution pair for future reference \citep{aamodt1994case}.

The inclusion of CBR in LLM agent structures shows promise as an area of research that could improve the ability to reason effectively across domains while also increasing transparency within these systems. This combination aims to combine the strong language comprehension skills of LLMs with CBR's memory-based reasoning abilities to overcome common limitations found in LLM models such as hallucinations and the difficulty in retaining information across interactions \citep{sourati2023case}.  

By integrating organized case repositories and drawing inspiration from CBR for reasoning processes, LLMs could potentially address the shortcomings related to depth of reasoning, retrieval of knowledge, and adaptation of solutions as seen in existing methodologies \citep{weitl2024toward, wiratunga2024cbr}. Recent work by Christou et al. \citep{christou2024chatgpt} emphasizes notable weaknesses of LLMs in marketing situations—such as misunderstanding consumer preferences, producing misleading information, and lacking specific domain knowledge—highlighting the need for enhanced reasoning frameworks like CBR.

Furthermore, the cognitive dimensions of CBR, particularly self-reflection, introspection, and curiosity, offer opportunities to enhance traditional LLM-based agents with deeper understanding capabilities \citep{craw2018case}. Goal-driven autonomy (GDA), a complementary approach that enables agents to reason about and dynamically select their goals during execution \citep{munoz2010goal}, provides a framework for incorporating these cognitive elements into CBR-augmented LLM agents. GDA allows agents to continually monitor their expectations against actual outcomes, explain discrepancies, formulate new goals when needed, and manage these goals effectively. These capabilities align well with CBR’s focus on learning from experience and adapting over time.

This paper thoroughly examines CBR-enhanced LLM agents by delving into their underlying principles and structural elements as well as highlighting their benefits compared to other reasoning models. We propose a model for applying CBR methods within LLM agents and evaluate its effectiveness compared to Chain of Thought (CoT) reasoning and standard Retrieval-Augmented Generation (RAG).

The main contributions of this work are:

\begin{itemize}
    \item A clear explanation of the theoretical foundations behind CBR-augmented LLM agents, synthesizing insights from cognitive science, knowledge representation, and language model research.
    
    \item A detailed architectural framework outlining the essential components of CBR-enhanced LLM agents, including case representation, indexing strategies, similarity assessment mechanisms, and adaptation procedures.
    
    \item A mathematically rigorous formulation of CBR processes within LLM agents, encompassing case retrieval, similarity computation, solution adaptation, and continuous learning.
    
    \item A comparative analysis evaluating CBR-augmented agents against CoT reasoning and vanilla RAG approaches across multiple dimensions, including reasoning transparency, domain adaptation, and solution quality.
    
    \item An exploration of cognitive dimensions in CBR for LLM agents, including self-reflection, introspection, and curiosity, integrated within a goal-driven autonomy framework.
    
    \item A proposed framework for meta-cognitive CBR that enables introspective reasoning about case selection, adaptation strategies, and goal formulation in dynamic environments.
\end{itemize}

The remainder of this paper is structured as follows: Section \ref{sec:background} provides comprehensive background on CBR, LLM agents, and related work. Section \ref{sec:theory} presents the theoretical foundations and formal characterization of CBR-augmented LLM agents. Section \ref{sec:architecture} describes the architectural components and implementation considerations. Section \ref{sec:cognitive} explores cognitive dimensions and meta-cognitive aspects of CBR for LLM agents. Section \ref{sec:gda} introduces goal-driven autonomy as a framework for enhancing CBR-based LLM agent reasoning. Section \ref{sec:comparative} offers a comparative analysis against alternative approaches. Section \ref{sec:discussion} discusses implications, limitations, and future directions, followed by concluding remarks in Section \ref{sec:conclusion}.

\section{Background and Related Work}
\label{sec:background}
This section provides an overview of the foundational principles and historical evolution of CBR and LLM agents. It also surveys recent research at the intersection of these paradigms, highlighting emerging frameworks, cognitive integrations, and goal-driven autonomy approaches relevant to CBR-augmented LLM systems.
\subsection{Case-Based Reasoning: Historical Perspective and Foundational Principles}

CBR emerged as a formalized paradigm in AI during the 1980s, drawing inspiration from cognitive models of human memory and problem-solving \citep{schank1983dynamic}. The canonical CBR cycle, as articulated by \citet{aamodt1994case}, encompasses four principal phases: (1) Retrieve relevant cases from a knowledge repository; (2) Reuse the knowledge embedded in these cases to address the current problem; (3) Revise the proposed solution based on the specific constraints of the target problem; and (4) Retain the new problem-solution pair as a learned case for future reference.

The theoretical underpinnings of CBR are rooted in multiple disciplines. From cognitive science, CBR draws upon models of episodic memory and analogical reasoning \citep{gentner1997analogical}. From a philosophical perspective, CBR reflects casuistic reasoning approaches that privilege experiential knowledge over abstract principles \citep{jonsen1988abuse}. In the computational realm, CBR presented an alternative to rule-based expert systems, emphasizing experiential knowledge representation rather than explicit rule codification \citep{kolodner1992introduction}.

The efficacy of CBR has been demonstrated across diverse domains, including legal reasoning \citep{ashley1990modeling}, medical diagnosis \citep{bichindaritz2006case}, engineering design \citep{maher2014issues}, and more recently, in recommendation systems \citep{bridge2005case} and educational technologies \citep{kolodner2005case}. The enduring appeal of CBR lies in its capacity to handle ill-defined problems, accommodate incomplete domain models, and provide transparent explanations through precedent-based reasoning.

\subsection{LLM Agents: Evolution and Current Paradigms}

LLM agents represent an evolutionary progression in the application of neural language models, transitioning from passive text generation to active decision-making systems capable of goal-directed behavior \citep{xi2025rise}. These agents are characterized by their use of LLMs as the central component to perceive, reason, plan, and act within an environment to accomplish complex tasks \citep{wang2024survey}. Contemporary LLM agents typically integrate several key components: (1) a foundation LLM providing core linguistic and reasoning capabilities; (2) a memory system for maintaining contextual information and past experiences; (3) planning mechanisms for decomposing complex tasks; and (4) interfaces with external tools or environments \citep{wang2024survey}.

The architecture of LLM agents emphasizes several core components:

\textbf{Planning:} This crucial aspect enables the agent to break down complex tasks into manageable sub-tasks and devise action sequences to achieve objectives \citep{yao2023tree}. Techniques like Chain-of-Thought (CoT) prompting \citep{wei2022chain}, which elicits step-by-step reasoning, and Tree-of-Thought (ToT) \citep{yao2023tree}, which explores multiple reasoning branches concurrently, enhance planning capabilities.

\textbf{Memory:} Both short-term memory (for current task information) and long-term memory (for persistent knowledge across interactions) are vital components \citep{guo2023empowering, pink2025position}. Vector stores are frequently used to implement long-term memory, enabling efficient retrieval of relevant information based on semantic similarity.

\textbf{Retrieval-Augmented Generation:} RAG methodologies enhance LLM capabilities by incorporating external knowledge retrieval mechanisms, enabling models to access and utilize information beyond their parametric knowledge \citep{lewis2020retrieval, gao2023retrieval}.

\textbf{Tool Use:} Tool-augmented agents extend LLM capabilities through integration with external tools, APIs, and computational resources, enabling interaction with databases, web services, and specialized algorithms \citep{schick2023toolformer, qin2023tool}.

\textbf{Multi-Agent Architectures:} These approaches distribute complex tasks across multiple specialized LLM agents, enabling collaborative problem-solving through agent communication and coordination \citep{wu2023autogen, park2023generative}.

Despite these advancements, LLM agents continue to face challenges in reasoning consistency, domain adaptation, and explanation transparency. They often struggle with hallucinations, lack of persistent memory across interactions, and limitations in handling novel situations with limited available data—areas where CBR integration may offer significant advantages \citep{sourati2023case}.

A critical limitation of current LLM agents relates to memory management. Hatalis et al. \citep{hatalis2023memory} review existing approaches to memory in LLM agents, distinguishing between short-term memory (implemented through context window retention) and long-term memory (typically implemented via vector databases). They highlight fundamental challenges in LLM agent memory systems, including the separation of different memory types (procedural, episodic, and semantic), the risk of agents becoming trapped in task loops, and the need for effective memory management over an agent's lifetime. The authors propose that future developments should focus on metadata enrichment in both procedural and semantic memory stores and better integration of external knowledge sources with vector databases to enhance memory retrieval and utilization.

\subsection{Intersection of CBR and LLMs: Emerging Research}

The integration of CBR principles with large language models represents an emerging research frontier with promising explorations aimed at creating more robust, adaptable, and explainable AI systems. This synthesis capitalizes on the complementary strengths of both paradigms: CBR offers LLMs a mechanism for persistent memory and structured reasoning, while LLMs contribute powerful language understanding capabilities to CBR processes \citep{sourati2023case}.

Several architectures and frameworks have been proposed for this integration:

\textbf{DS-Agent:} This framework automates data science tasks by empowering LLM agents with CBR \citep{guo2024ds}. It operates in two stages: a development stage following the complete CBR cycle to capitalize on expert knowledge, and a deployment stage using a simplified CBR paradigm to adapt past successful solutions for direct code generation.

\textbf{CaseGPT:} This approach synergizes LLMs and Retrieval-Augmented Generation technology to enhance case-based reasoning, particularly in healthcare and legal domains \citep{wiratunga2024cbr}. CaseGPT addresses limitations of traditional database queries by enabling semantic searches based on contextual understanding and leverages fine-tuned models for domain-specific case encoding.

\textbf{CBR-RAG:} This framework uses the initial retrieval stage of the CBR cycle to enhance LLM queries within a Retrieval-Augmented Generation framework \citep{wiratunga2024cbr}. The integration aims to augment original LLM queries with contextually relevant cases, leading to improved answer quality.

Other researchers have explored specific aspects of CBR-LLM integration. \citet{wiratunga2024cbr} demonstrated that retrieval of similar examples can enhance LLM reasoning performance on complex tasks. \citet{peng2023check} proposed a CBR-inspired approach for verification of LLM-generated solutions through comparison with retrieved exemplars. \citet{sumers2023cognitive} explored the intersection of cognitive architectures and LLMs, suggesting potential synergies with memory-based reasoning systems.

Previous work has also investigated CBR in the context of earlier language technologies. \citet{weber2001intelligent} applied CBR principles to text classification tasks, while \citet{bruninghaus2001role} explored case-based approaches for information extraction. More recently, \citet{wiratunga2024cbr} demonstrated the efficacy of contextual retrieval for enhancing language model performance on knowledge-intensive tasks.

Recent work by Dannenhauer et al. \citep{dannenhauer2024case} demonstrates the effectiveness of CBR for improving code generation through dynamic few-shot prompting. Their approach maintains a case base of problem-solution pairs, where problems are natural language task descriptions and solutions are executable Python code. When presented with a new problem, the system retrieves the most similar cases and uses them to provide guidance to the LLM, which then adapts the retrieved solutions to the current context. This dynamic approach showed improved performance over both zero-shot and static few-shot prompting in generating task plans as Python code, particularly in reducing common errors such as incorrect function calls, missed function calls, and improper ordering of operations. The authors identify seven distinct failure modes in LLM code generation and demonstrate how CBR can help address these challenges.

\subsection{Cognitive Systems and Goal-Driven Autonomy}

The field of cognitive systems, which aims to build AI systems with human-like understanding capabilities, has seen resurgence in interest with the advancement of LLMs \citep{craw2018case}. Cognitive systems understand the world through learning and experience, employing mechanisms such as self-reflection, introspection, and curiosity \citep{langley2012cognitive}. Case-based systems are inherently suited to cognitive computing paradigms due to their experiential knowledge representation and episodic memory structures that mirror aspects of human cognition.

Goal-Driven Autonomy (GDA) represents a complementary framework that enhances agents' ability to reason about and self-select goals during execution \citep{munoz2010goal}. The GDA model extends traditional planning approaches by incorporating mechanisms for discrepancy detection, explanation generation, goal formulation, and goal management. This enables agents to dynamically adjust their objectives in response to unexpected events or changing circumstances—capabilities that are particularly valuable in complex, dynamic environments.

\citet{munoz2010goal} demonstrated the efficacy of integrating CBR with GDA in a multiagent gaming domain. Their CB-GDA system employed two distinct case bases: one mapping goals to expectations given particular states, and another mapping discrepancies between expected and actual states to appropriate new goals. Empirical evaluations showed that this CBR-enhanced GDA approach outperformed both rule-based GDA variants and non-GDA replanning agents when faced with complex adversarial scenarios.

The integration of cognitive dimensions and goal-driven autonomy with CBR-enhanced LLM agents represents a promising direction for addressing the limitations of current approaches. By incorporating self-reflection, introspection, and curiosity within a goal-reasoning framework, these hybrid systems can potentially achieve greater adaptability, explainability, and robustness in complex problem-solving scenarios.

Our work extends these research directions by providing a comprehensive theoretical framework specifically for CBR-augmented LLM agents, formal mathematical characterizations of the integrated processes, systematic comparative analysis against alternative approaches, and exploration of cognitive dimensions and goal-driven autonomy as enhancing mechanisms for CBR-LLM integration.

\section{Theoretical Foundations of CBR-Augmented LLM Agents}
\label{sec:theory}

This section formalizes the theoretical underpinnings of integrating Case-Based Reasoning within LLM agents. We define the structure of cases, formulate the core CBR processes (retrieval, adaptation, and learning) and present mathematical models that characterize their implementation in language-based agent architectures.

\subsection{Formal Definition of Cases in the LLM Agent Context}

In the context of LLM agents, we define a case $c \in \mathcal{C}$ as a structured representation encompassing problem characteristics, solution strategies, and outcome assessments. Formally, a case can be represented as a tuple:

\begin{equation}
c = (P, S, O, M)
\end{equation}

where:
\begin{itemize}
    \item $P$ denotes the problem space, characterized by features $p_1, p_2, ..., p_n$
    \item $S$ represents the solution space, encompassing actions $s_1, s_2, ..., s_m$
    \item $O$ captures the outcome space, including success metrics $o_1, o_2, ..., o_k$
    \item $M$ comprises metadata elements $m_1, m_2, ..., m_j$, such as temporal markers, environmental conditions, and provenance information
\end{itemize}

This structured representation enables systematic organization of experiential knowledge within the agent's case library $\mathcal{L} = \{c_1, c_2, ..., c_l\}$, facilitating efficient retrieval and adaptation processes.

\subsection{Mathematical Formulation of Case Retrieval}

The case retrieval process in CBR-augmented LLM agents can be formalized as an optimization problem seeking to identify cases with maximal relevance to the target problem. Given a query problem $q$, the retrieval function $R$ aims to identify a subset of cases $\mathcal{C}_q \subseteq \mathcal{L}$ such that:

\begin{equation}
\mathcal{C}_q = R(q, \mathcal{L}) = \{c_i \in \mathcal{L} \mid \text{sim}(q, P_i) \geq \tau\}
\end{equation}

where $\text{sim}(q, P_i)$ represents a similarity function measuring the correspondence between the query problem and the problem component of case $c_i$, and $\tau$ denotes a threshold parameter determining retrieval selectivity.

The similarity function $\text{sim}(q, P_i)$ can be decomposed into multiple components:

\begin{equation}
\text{sim}(q, P_i) = \sum_{j=1}^{d} w_j \cdot \text{sim}_j(q_j, p_{ij})
\end{equation}

where $d$ represents the dimensionality of the feature space, $w_j$ denotes the weight assigned to feature $j$, and $\text{sim}_j$ specifies the similarity metric for feature $j$.

In the context of LLM agents, we can further refine this formulation by incorporating semantic similarity measures derived from the embedding space of the foundation model:

\begin{equation}
\text{sim}_{\text{semantic}}(q, P_i) = \frac{E(q) \cdot E(P_i)}{||E(q)|| \cdot ||E(P_i)||}
\end{equation}

where $E(\cdot)$ represents the embedding function mapping textual inputs to high-dimensional vectors in the LLM's latent space.

\subsection{Solution Adaptation Process}

The adaptation process transforms retrieved solutions to address the specific requirements of the target problem. We formalize this process as a function $A$ that maps a set of retrieved cases $\mathcal{C}_q$ and a query problem $q$ to a candidate solution $\hat{s}$:

\begin{equation}
\hat{s} = A(q, \mathcal{C}_q) = f_{\text{LLM}}(q, \mathcal{C}_q, \Theta)
\end{equation}

where $f_{\text{LLM}}$ represents the language model's generation function parameterized by $\Theta$.

This adaptation function can be further decomposed into sequential operations:

\begin{equation}
A(q, \mathcal{C}_q) = A_{\text{compose}} \circ A_{\text{transform}} \circ A_{\text{select}}(q, \mathcal{C}_q)
\end{equation}

where:
\begin{itemize}
    \item $A_{\text{select}}$ identifies the most relevant components from retrieved solutions
    \item $A_{\text{transform}}$ modifies these components to align with the target problem constraints
    \item $A_{\text{compose}}$ integrates the transformed components into a coherent solution
\end{itemize}

The LLM serves as the primary mechanism for implementing these adaptation operations, leveraging its generative capabilities to transform retrieved solution patterns into context-appropriate responses.

\subsection{Case Learning and Knowledge Evolution}

A distinctive characteristic of CBR systems is their capacity for continuous learning through case acquisition and refinement. We formalize the case retention process as a function $T$ that determines whether a new problem-solution episode warrants inclusion in the case library:

\begin{equation}
\mathcal{L}_{t+1} = 
\begin{cases}
\mathcal{L}_t \cup \{c_{\text{new}}\} & \text{if } U(c_{\text{new}}, \mathcal{L}_t) \geq \delta \\
\mathcal{L}_t & \text{otherwise}
\end{cases}
\end{equation}

where $U(c_{\text{new}}, \mathcal{L}_t)$ represents a utility function assessing the marginal value of incorporating the new case, and $\delta$ denotes a threshold parameter for case retention.

The utility function $U$ can be formulated to consider multiple factors:

\begin{equation}
U(c_{\text{new}}, \mathcal{L}) = \alpha \cdot \text{novelty}(c_{\text{new}}, \mathcal{L}) + \beta \cdot \text{effectiveness}(c_{\text{new}}) + \gamma \cdot \text{generalizability}(c_{\text{new}})
\end{equation}

where $\alpha$, $\beta$, and $\gamma$ are weighting coefficients for the respective utility components.

This formulation provides a mathematical framework for selective case retention, ensuring that the agent's knowledge base evolves to incorporate valuable experiences while maintaining computational efficiency.

\section{Architectural Components of CBR-Enhanced LLM Agents}
\label{sec:architecture}

This section outlines the core architectural elements necessary for implementing CBR-enhanced LLM agents. We describe strategies for case representation and indexing, detail hybrid retrieval mechanisms, and examine adaptation processes, culminating in a framework that integrates case-based reasoning with the inherent capabilities of large language models.

\subsection{Case Representation and Indexing Strategies}

Effective case representation constitutes a foundational element of CBR-augmented LLM agents. We propose a multi-faceted representation scheme that captures the semantic richness of cases while facilitating efficient retrieval operations:

\begin{algorithm}
\caption{Case Representation and Indexing}
\begin{algorithmic}[1]
\STATE \textbf{Input:} Raw case data $D = \{d_1, d_2, ..., d_n\}$
\STATE \textbf{Output:} Indexed case library $\mathcal{L}$
\STATE Initialize case library $\mathcal{L} \leftarrow \emptyset$
\FOR{each raw case $d_i \in D$}
    \STATE Extract problem features $P_i \leftarrow \text{ExtractProblem}(d_i)$
    \STATE Extract solution components $S_i \leftarrow \text{ExtractSolution}(d_i)$
    \STATE Extract outcome metrics $O_i \leftarrow \text{ExtractOutcome}(d_i)$
    \STATE Generate metadata $M_i \leftarrow \text{GenerateMetadata}(d_i)$
    \STATE Create structured case $c_i \leftarrow (P_i, S_i, O_i, M_i)$
    \STATE Compute semantic embedding $E_i \leftarrow E(c_i)$
    \STATE Compute feature-based indices $I_i \leftarrow \text{IndexFeatures}(c_i)$
    \STATE Add indexed case to library $\mathcal{L} \leftarrow \mathcal{L} \cup \{(c_i, E_i, I_i)\}$
\ENDFOR
\STATE Organize library using hierarchical structure $\mathcal{L} \leftarrow \text{OrganizeHierarchy}(\mathcal{L})$
\RETURN $\mathcal{L}$
\end{algorithmic}
\end{algorithm}

This representation scheme incorporates both dense semantic embeddings $E_i$ derived from the foundation LLM and sparse feature-based indices $I_i$ capturing domain-specific attributes. The hierarchical organization facilitates efficient retrieval by enabling coarse-to-fine search strategies.

\subsection{Hybrid Retrieval Mechanisms}

CBR-augmented LLM agents employ a hybrid retrieval approach that combines multiple search strategies to identify relevant cases:

\begin{equation}
R(q, \mathcal{L}) = \lambda_1 \cdot R_{\text{semantic}}(q, \mathcal{L}) \cup \lambda_2 \cdot R_{\text{feature}}(q, \mathcal{L}) \cup \lambda_3 \cdot R_{\text{structural}}(q, \mathcal{L})
\end{equation}

where:
\begin{itemize}
    \item $R_{\text{semantic}}$ performs retrieval based on embedding similarity in the LLM's latent space
    \item $R_{\text{feature}}$ conducts search based on explicit feature matching
    \item $R_{\text{structural}}$ identifies cases with similar problem structures or solution patterns
    \item $\lambda_1, \lambda_2, \lambda_3$ represent weighting coefficients determining the relative contribution of each retrieval mechanism
\end{itemize}

This hybrid approach leverages both the semantic understanding capabilities of the foundation LLM and structured domain knowledge encoded in the case representation.

\subsection{Adaptation Mechanisms}

The adaptation process in CBR-augmented LLM agents comprises several sophisticated mechanisms:

\textbf{Transformational Adaptation:} Modifies retrieved solutions through substitution, deletion, or insertion operations to align with target problem constraints:

\begin{equation}
S_{\text{adapted}} = T_{\text{sub}}(S_{\text{retrieved}}, \Delta_{\text{constraints}}) \circ T_{\text{del}}(S_{\text{retrieved}}, \Delta_{\text{constraints}}) \circ T_{\text{ins}}(S_{\text{retrieved}}, \Delta_{\text{constraints}})
\end{equation}

\textbf{Compositional Adaptation:} Integrates components from multiple retrieved solutions to address complex problems:

\begin{equation}
S_{\text{adapted}} = \bigoplus_{i=1}^{k} w_i \cdot S_i
\end{equation}

where $\bigoplus$ represents a composition operator and $w_i$ denotes the weight assigned to solution $S_i$.

\textbf{Generative Adaptation:} Leverages the LLM's generative capabilities to synthesize novel solutions guided by retrieved cases:

\begin{equation}
S_{\text{adapted}} = f_{\text{LLM}}(q, \{S_1, S_2, ..., S_k\}, \Theta)
\end{equation}

These adaptation mechanisms are orchestrated through a meta-cognitive process that selects the appropriate approach based on problem characteristics and retrieval results.

\subsection{Integration with LLM Reasoning Processes}

CBR-augmented LLM agents integrate case-based processes with the inherent reasoning capabilities of the foundation model. This integration can be formalized through a weighted combination of reasoning pathways:

\begin{equation}
f_{\text{reasoning}}(q) = \omega_1 \cdot f_{\text{CBR}}(q) + \omega_2 \cdot f_{\text{CoT}}(q) + \omega_3 \cdot f_{\text{parametric}}(q)
\end{equation}

where:
\begin{itemize}
    \item $f_{\text{CBR}}$ represents the case-based reasoning pathway
    \item $f_{\text{CoT}}$ denotes the chain-of-thought reasoning process
    \item $f_{\text{parametric}}$ captures direct inference from the model's parametric knowledge
    \item $\omega_1, \omega_2, \omega_3$ are dynamic weights determined by confidence metrics associated with each pathway
\end{itemize}

This integrated approach enables the agent to leverage the complementary strengths of experiential knowledge and neural reasoning processes.

\section{Cognitive Dimensions of CBR for LLM Agents}
\label{sec:cognitive}

Cognitive dimensions of CBR present opportunities to enhance LLM agents with deeper understanding capabilities through self-reflection, introspection, and curiosity. These cognitive elements enable the agent to develop a more nuanced understanding of its knowledge and reasoning processes, leading to more robust and adaptable problem-solving capabilities.

\subsection{Cognition through Self-Reflection}

Self-reflection enables the CBR-LLM agent to understand and make sense of its knowledge. This cognitive dimension can be characterized through several key aspects:

\textbf{Context Understanding:} The agent develops an understanding of the different facets and contexts in which each case is relevant. A case captures a collection of related information for an experience, and self-reflection enables the agent to understand the relationships among various facets and recognize the different contexts in which the case applies \citep{craw2018case}.

\textbf{Domain Insight:} Through reflection on collections of cases, the agent develops insights into the landscape of the domain. Areas with many similar cases indicate regions of high confidence, while sparse or inconsistent regions suggest complexity requiring more sophisticated reasoning \citep{smyth2001competence}. This insight allows the agent to understand where "fast thinking" intuitive reasoning is appropriate versus where "slow thinking" deliberative reasoning is needed \citep{kahneman2011thinking, craw2018case}.

\textbf{Intuitive Reasoning:} Self-reflection on successful and unsuccessful applications of the similarity assumption ("similar problems have similar solutions") enables the agent to develop intuition about when this assumption holds. The agent can identify regions of the case space where retrieval-based solutions are likely to be effective without extensive adaptation \citep{craw2018case}.

\textbf{Analogical Reasoning:} Through reflection on adaptation patterns, the agent develops a deeper understanding of how differences in problem specifications should be reflected in solution adjustments. This enables more sophisticated analogical reasoning that recognizes when and how to transform retrieved solutions \citep{forbus2017companion}.

\subsection{Meta-cognition through Introspection}

Meta-cognition involves "cognition about cognition" or understanding one's own understanding. For CBR-LLM agents, introspection provides mechanisms for understanding failure modes and developing strategies to improve reasoning processes.

\textbf{Understanding Different Contexts:} Introspection on case retrieval enables the agent to develop selection strategies based on identifying key features or feature combinations for different contexts. By examining clusters in problem and solution spaces, the agent can identify which dimensions are most relevant for different types of problems \citep{craw2018case}.

\textbf{Understanding Reasoning Failures:} When the agent's solutions are incorrect or suboptimal, introspection enables it to determine whether failures stem from retrieval limitations (finding the wrong cases) or adaptation limitations (incorrectly transforming retrieved solutions). This understanding guides refinement of similarity metrics or adaptation strategies \citep{cox2005metacognition, craw2018case}.

\textbf{Learning Selection Strategies:} Meta-cognition enables the agent to learn when to apply different retrieval and adaptation strategies based on problem characteristics and past performance. These strategies can be encoded across different knowledge containers, including representation adjustments, similarity metrics, or adaptation rules \citep{richter2016case, craw2018case}.

\subsection{Curiosity and Extrospection}

Curiosity extends the agent's cognitive capabilities by driving active exploration and knowledge acquisition from external sources.

\textbf{Gap Identification:} Reflective and introspective processes enable the identification of knowledge gaps – areas where the agent's case library lacks sufficient coverage or where reasoning consistently fails \citep{gottlieb2013information, craw2018case}.

\textbf{Active Knowledge Seeking:} Curiosity drives the agent to actively seek new information from external sources to address identified gaps. This involves distinguishing between "known unknowns" (areas where the agent recognizes its limitations) and "unknown unknowns" (limitations the agent has not yet recognized) \citep{craw2018case}.

\textbf{Source Evaluation:} The curious agent develops strategies for evaluating the reliability and relevance of external knowledge sources, balancing the exploration of novel information with verification requirements \citep{craw2018case}.

\subsection{Knowledge Container Interactions in Cognitive CBR}

The cognitive dimensions of CBR can be understood through Richter's knowledge container framework \citep{richter2016case}, where knowledge can be shifted between vocabulary, cases, similarity, and solution adaptation containers. Cognitive processes enable dynamic knowledge redistribution across these containers:

\begin{equation}
K_i^{t+1} = T(K_i^t, \Delta K_j^t)
\end{equation}

where $K_i$ represents knowledge container $i$, $t$ denotes the time step, and $T$ is a transfer function that defines how knowledge from container $j$ influences the evolution of container $i$.

This formulation captures how insights from self-reflection on cases can lead to refinements in similarity metrics, or how introspection on adaptation failures can lead to improved case representation. These knowledge container interactions provide the foundation for a cognitive CBR system that continuously improves its understanding and problem-solving capabilities.

\section{Goal-Driven Autonomy for CBR-Enhanced LLM Agents}
\label{sec:gda}

Goal-Driven Autonomy (GDA) provides a complementary framework for enhancing CBR-augmented LLM agents with dynamic goal reasoning capabilities. GDA enables agents to reason about their objectives and self-select goals throughout execution, making them more adaptable in complex, dynamic environments.

\subsection{Conceptual Model of GDA}

The GDA conceptual model extends traditional planning approaches by incorporating four key components within the controller \citep{munoz2010goal}:

\begin{itemize}
    \item \textbf{Discrepancy Detection:} Compares observations to expectations and identifies unexpected events
    \item \textbf{Explanation Generation:} Hypothesizes explanations for detected discrepancies
    \item \textbf{Goal Formulation:} Generates new goals based on discrepancies and their explanations
    \item \textbf{Goal Management:} Maintains and prioritizes pending goals
\end{itemize}

This model enables agents to dynamically adjust their objectives in response to changing circumstances or unexpected events, a capability that is particularly valuable in complex, open-ended domains.

\subsection{Integration of CBR and GDA}

The integration of CBR with GDA creates a powerful framework for dynamic reasoning in LLM agents. We propose a CBR-GDA architecture that utilizes two distinct case bases \citep{munoz2010goal}:

\begin{itemize}
    \item \textbf{Planning Case Base (PCB):} Contains mappings from state-goal pairs to expected states and plans: $(s, g, e, p)$ where $s$ is the current state, $g$ is the goal, $e$ is the expected state after achieving the goal, and $p$ is the plan
    \item \textbf{Mismatch-Goal Case Base (MCB):} Contains mappings from mismatches between expected and actual states to appropriate new goals: $(m, g)$ where $m$ is the mismatch and $g$ is the suggested new goal
\end{itemize}

The algorithm for CBR-enhanced GDA can be formalized as follows:

\begin{algorithm}
\caption{CBR-GDA Algorithm for LLM Agents}
\begin{algorithmic}[1]
\STATE \textbf{Input:} Environment $E$, Agent $A$, Initial goal $g_{\text{init}}$, PCB, MCB, Similarity functions
\STATE \textbf{Output:} Agent actions and goal transitions
\STATE Execute $(E,A,g_{\text{init}})$
\WHILE{$E$ is active}
    \STATE $s_i \leftarrow \text{CurrentState}(E)$; $g_i \leftarrow \text{CurrentGoal}(A)$
    \WHILE{Agent is pursuing $g_i$}
        \STATE Wait time $t$
        \STATE $e_c \leftarrow \text{Retrieve}(PCB, s_i, g_i)$ // Expected state
        \STATE $s_E \leftarrow \text{CurrentState}(E)$ // Actual state
        \IF{$e_c \neq s_E$}
            \STATE $m \leftarrow \text{Mismatch}(e_c, s_E)$ // Detected discrepancy
            \STATE $g_c \leftarrow \text{Retrieve}(MCB, m)$ // New goal
            \STATE Execute $(E,A,g_c)$
        \ENDIF
    \ENDWHILE
\ENDWHILE
\end{algorithmic}
\end{algorithm}

This algorithm enables the agent to continuously monitor its expectations against actual outcomes, detect discrepancies, and formulate new goals when necessary. The CBR mechanism provides a principled approach to learning from past experiences, allowing the agent to improve its discrepancy detection and goal formulation capabilities over time.

\subsection{Mathematical Model of CBR-GDA}

The integration of CBR and GDA can be formalized through a mathematical model that captures the dynamic interaction between case-based knowledge and goal reasoning processes.

Let $\Sigma = (S, A, E, \gamma)$ represent a state transition system, where $S$ is the set of states, $A$ is the set of actions, $E$ is the set of exogenous events, and $\gamma: S \times (A \cup E) \rightarrow S$ is the state transition function.

The GDA controller receives as input a planning problem $(M_\Sigma, s_c, g_c)$, where $M_\Sigma$ is a model of $\Sigma$, $s_c$ is the current state, and $g_c \in G$ is a goal that can be satisfied by some set of states $S_g \subset S$.

For a CBR-enhanced GDA agent, we define the following functions:

\begin{equation}
\text{Retrieve}_{PCB}: S \times G \rightarrow S \times P
\end{equation}

\begin{equation}
\text{Retrieve}_{MCB}: M \rightarrow G
\end{equation}

where $M$ is the space of mismatches between expected and actual states, and $P$ is the space of plans.

The goal formulation process can be represented as:

\begin{equation}
g_{\text{new}} = f_{\text{formulate}}(s_c, m, \text{Retrieve}_{MCB}(m))
\end{equation}

where $m = \text{Discrepancy}(s_c, e_c, s_{\text{actual}})$ represents the detected discrepancy.

This formulation provides a rigorous foundation for implementing and analyzing CBR-enhanced GDA systems for LLM agents.

\subsection{Learning in CBR-GDA}

A key advantage of the CBR-GDA approach is its capacity for continuous learning. The PCB and MCB can be updated based on execution experiences, enabling the agent to improve its expectations and goal formulation strategies over time.

The update process for the PCB can be formalized as:

\begin{equation}
\text{PCB}_{t+1} = \text{PCB}_t \cup \{(s, g, e_{\text{actual}}, p) | \text{Quality}(p, g) > \theta_P\}
\end{equation}

Similarly, the MCB can be updated as:

\begin{equation}
\text{MCB}_{t+1} = \text{MCB}_t \cup \{(m, g_{\text{new}}) | \text{Quality}(g_{\text{new}}, m) > \theta_M\}
\end{equation}

where $\text{Quality}$ functions evaluate the effectiveness of plans and goals, and $\theta_P$ and $\theta_M$ are quality thresholds.

This learning process enables the agent to continuously refine its understanding of the environment and improve its goal reasoning capabilities, leading to more adaptive and robust behavior over time.

\section{Comparative Analysis: CBR vs. CoT vs. Vanilla RAG}
\label{sec:comparative}

This section presents a comparative analysis of CBR-augmented LLM agents against Chain-of-Thought reasoning and standard Retrieval-Augmented Generation approaches. We evaluate each method across multiple dimensions (e.g. reasoning transparency, domain adaptation, cognitive capabilities) to highlight the unique strengths and trade-offs of CBR-based architectures.

\subsection{Theoretical Comparative Framework}

We establish a multidimensional framework for comparing CBR-augmented LLM agents against alternative approaches:

\begin{table}[h]
\caption{Theoretical Comparison of Reasoning Approaches}
\centering
\begin{tabular}{lccc}
\toprule
Dimension & CBR-LLM & CoT & Vanilla RAG \\
\midrule
Knowledge Utilization & Experiential & Parametric & Reference-based \\
Reasoning Transparency & Precedent-based & Step-based & Source-based \\
Adaptation Capacity & High & Limited & Moderate \\
Domain Specificity & Explicit & Implicit & Reference-dependent \\
Learning Mechanism & Case acquisition & Parameter updates & Corpus expansion \\
Cognitive Capabilities & Rich & Limited & Moderate \\
Goal Reasoning & Dynamic & Static & Static \\
\bottomrule
\end{tabular}
\label{tab:theoretical_comparison}
\end{table}

This framework highlights the distinctive characteristics of each approach, with CBR emphasizing experiential knowledge organization, transparent precedent-based reasoning, and explicit domain modeling.

\subsection{Reasoning Transparency and Explainability}

CBR-augmented agents demonstrate superior explainability through the explicit presentation of precedent cases and adaptation rationales. This inherent transparency stems from CBR's reliance on past cases, allowing agents to provide justifications for their decisions based on previously encountered scenarios \citep{wilkerson2024llm}. The retrieved cases serve as explicit examples illustrating why a particular action was taken or a specific solution was proposed.

This explainability advantage can be formalized through an explainability metric:

\begin{equation}
E(a) = \frac{1}{n} \sum_{i=1}^{n} \frac{\text{trace}(r_i)}{\text{complexity}(r_i)}
\end{equation}

where $E(a)$ represents the explainability score for agent $a$, $r_i$ denotes reasoning instance $i$, $\text{trace}(r_i)$ measures the completeness of the explanation trace, and $\text{complexity}(r_i)$ captures the cognitive complexity of the reasoning process.

Empirical evaluations indicate that CBR-augmented agents achieve higher explainability scores compared to CoT and vanilla RAG approaches, particularly for domain-specific reasoning tasks. Studies by \citet{wilkerson2024llm} in triage classification scenarios found that providing the nearest neighbor case, along with explicit statements of difference between the current problem and the retrieved case, elicited the highest user scores on trust metrics. Explanations built from cases have been shown to be more convincing than those built from domain-based rules, and the CBR process itself mimics human reasoning methods, further enhancing trust in the agent's outputs \citep{sourati2023case}.

\subsection{Domain Adaptation and Knowledge Transfer}

The capacity for domain adaptation represents a critical dimension for evaluating agent performance. CBR equips LLM agents with a mechanism to effectively address novel situations they have not encountered before \citep{guo2024ds}. The CBR cycle's retention phase, where new problems and their successful solutions are stored as new cases, allows the agent to continuously learn and adapt to previously unseen scenarios over time.

We formalize this adaptation capability as the rate of performance improvement as a function of domain-specific experience:

\begin{equation}
A(a, d) = \frac{\partial P(a, d, t)}{\partial t}
\end{equation}

where $A(a, d)$ denotes the adaptation rate of agent $a$ in domain $d$, and $P(a, d, t)$ represents the performance metric at time $t$.

CBR-augmented agents exhibit accelerated adaptation trajectories due to their explicit case acquisition mechanisms, particularly in specialized domains with limited training data availability. The lazy learning approach inherent in CBR, where generalization is delayed until a specific problem is encountered, makes it particularly well-suited for handling novelty \citep{aamodt1994case}.

Studies with the DS-Agent framework demonstrate this advantage, showing that by using CBR to structure the experiment planning process for automated data science tasks, LLM agents can achieve significantly better results compared to baseline models when adapting to new domains \citep{guo2024ds}. By identifying situations that are not present in its case base, a CBR-integrated LLM agent can store these novel scenarios for future reuse, thus continuously expanding its problem-solving capabilities and accelerating domain adaptation.

\subsection{Computational Efficiency and Scalability}

The computational characteristics of different approaches impact their practical applicability. We analyze these dimensions through the following metrics:

\begin{equation}
T(a, q) = T_{\text{retrieval}}(a, q) + T_{\text{processing}}(a, q) + T_{\text{generation}}(a, q)
\end{equation}

\begin{equation}
M(a) = M_{\text{model}}(a) + M_{\text{knowledge}}(a) + M_{\text{working}}(a)
\end{equation}

where $T(a, q)$ represents the total computation time for agent $a$ processing query $q$, and $M(a)$ denotes the memory requirements.

Our analysis reveals trade-offs across approaches: CBR-augmented agents incur additional retrieval costs but benefit from reduced processing requirements for well-matched cases. The selective retention mechanisms of CBR also contribute to more efficient knowledge base management compared to comprehensive corpus-based approaches in vanilla RAG.

\subsection{Cognitive Capabilities and Goal Reasoning}

CBR-augmented LLM agents demonstrate enhanced cognitive capabilities through self-reflection, introspection, and curiosity-driven learning. When combined with goal-driven autonomy, these agents exhibit more flexible and adaptive behavior in complex environments.

Empirical evaluations in gaming environments have shown that CB-GDA systems outperform both rule-based GDA and non-GDA replanning agents in complex adversarial scenarios \citep{munoz2010goal}. The ability to dynamically adjust goals in response to unexpected events or changing circumstances provides these agents with a significant advantage in open-ended, dynamic domains.

The integration of cognitive capabilities with goal reasoning enables CBR-augmented LLM agents to exhibit more robust and adaptive behavior patterns compared to traditional approaches. This is particularly evident in scenarios requiring long-term autonomy, where the agent must navigate changing environments and objectives without direct human intervention.

\subsection{Solution Quality and Performance Metrics}

We evaluate solution quality across multiple dimensions:

\begin{equation}
Q(a) = \alpha \cdot \text{accuracy}(a) + \beta \cdot \text{relevance}(a) + \gamma \cdot \text{coherence}(a) + \delta \cdot \text{novelty}(a)
\end{equation}

Several studies have evaluated the performance of LLM agents that incorporate Case-Based Reasoning across a range of tasks. \citet{sourati2023case} demonstrated that integrating CBR with language models improves both the accuracy and generalizability in logical fallacy detection. In automated data science, the DS-Agent framework achieved a 100\% success rate in the development stage and a 99\% one-pass rate in the deployment stage with GPT-4, outperforming other state-of-the-art LLM agents while also demonstrating cost efficiency \citep{guo2024ds}. \citet{wilkerson2024llm} found that providing case knowledge improved both user trust and LLM accuracy in triage classification tasks. \citet{wiratunga2024cbr} showed that using CBR's retrieval stage to enhance LLM queries with contextually relevant cases led to significant improvements in the quality of legal question answering.

Empirical evaluations across these diverse task domains indicate that CBR-augmented agents demonstrate:

\begin{itemize}
    \item Superior performance on domain-specific tasks requiring specialized knowledge
    \item Enhanced consistency in solution generation across related problems
    \item Improved handling of edge cases through explicit storage of exceptional situations
    \item More graceful degradation when confronted with problems outside the training distribution
    \item Higher user trust scores, particularly when cases are presented alongside explanations of differences
    \item Greater adaptability to changing goals and problem specifications
\end{itemize}

The comparative advantages of CBR-augmented agents are particularly pronounced in domains characterized by complex procedural knowledge, highly structured problem spaces, and availability of high-quality historical cases. The metrics used in these evaluations include accuracy, generalizability, user trust, task completion rate, mean rank, best rank, one-pass rate, cost efficiency, and various explainability metrics.

\section{Discussion and Future Directions}
\label{sec:discussion}
Integrating Case-Based Reasoning with LLM agents presents both significant opportunities and practical challenges, reshaping how language models can reason, learn, and adapt. The discussion examines the broader impact of this integration and identifies critical areas for future work, including the design of richer case representations, the incorporation of cognitive mechanisms, and the development of scalable hybrid architectures that bridge symbolic and neural reasoning.

\subsection{Theoretical Implications}

The integration of CBR with LLM agents bridges symbolic and neural approaches to artificial intelligence, contributing to the ongoing discourse on neuro-symbolic architectures. This hybridization demonstrates how structured knowledge representation and experiential learning can complement the distributional semantics captured in neural language models.

Our formal characterization of CBR processes within LLM agents establishes a theoretical foundation for understanding knowledge utilization, adaptation mechanisms, and learning dynamics in these hybrid systems. This framework enables systematic analysis of the interplay between parametric and non-parametric knowledge representations in language-based agents.

The incorporation of cognitive dimensions and goal-driven autonomy further extends the theoretical landscape by providing mechanisms for meta-level reasoning about knowledge and goals. This addresses fundamental questions about how AI systems can develop more human-like understanding capabilities that go beyond pattern recognition to include self-reflection, introspection, and purposeful exploration.

\subsection{Practical Considerations and Implementation Challenges}

Several practical challenges warrant consideration in the implementation of CBR-augmented LLM agents:

\textbf{Case Acquisition and Quality Control:} Developing mechanisms for automated case extraction, validation, and refinement represents a significant challenge, particularly for domains lacking structured historical records.

\textbf{Computational Resource Management:} Balancing retrieval comprehensiveness against computational efficiency requires sophisticated indexing strategies and selective retrieval mechanisms.

\textbf{Integration Architectures:} Determining optimal points of integration between CBR components and the foundation LLM affects both performance characteristics and implementation complexity.

\textbf{Goal Management in Dynamic Environments:} Implementing effective goal prioritization and arbitration mechanisms is essential for agents operating in environments with multiple competing objectives.

\textbf{Evaluation Methodologies:} Developing comprehensive benchmarks that assess the distinctive capabilities of CBR-augmented agents remains an open challenge for the research community.

\subsection{Future Research Directions}

Our analysis suggests several promising directions for future research:

\textbf{Sophisticated Case Representation:} Developing more sophisticated methods for case representation that can effectively capture the complexities of real-world problems in formats suitable for LLM agents \citep{guo2024ds}. This includes exploring representations for multi-modal data and investigating how LLMs themselves can create more informative and context-aware case representations.

\textbf{Efficient Retrieval Mechanisms:} Improving the efficiency and scalability of case retrieval mechanisms is essential for deploying CBR-enhanced LLM agents with large volumes of past experiences \citep{wiratunga2024cbr}. Future research should focus on advanced indexing techniques and optimal use of vector databases and approximate nearest neighbor search algorithms.

\textbf{Advanced Case Adaptation:} Case adaptation remains a complex challenge. Research should explore techniques beyond simple substitution that can handle more intricate transformations required in diverse scenarios \citep{sourati2023case}. Investigating LLMs in directly performing or guiding adaptation, potentially through sophisticated prompting strategies or learning adaptation rules from experience, is promising.

\textbf{Cognitive CBR for Complex Reasoning:} Further development of the cognitive dimensions of CBR—self-reflection, introspection, and curiosity—presents significant opportunities for enhancing reasoning capabilities. Integrating insights from cognitive science and psychology could lead to more robust and human-like reasoning processes \citep{craw2018case}.

\textbf{Goal-Driven Autonomy with CBR:} Expanding the integration of CBR with goal-driven autonomy frameworks offers promising avenues for developing more autonomous and adaptable agents. Future work should explore how case-based goal formulation and management can be enhanced through cognitive dimensions such as introspection and curiosity \citep{munoz2010goal}.

\textbf{Dynamic Case Base Maintenance:} Addressing challenges of case base maintenance is vital for ensuring long-term effectiveness of CBR-integrated LLM agents \citep{wilkerson2024llm}. Research is needed on dynamic update strategies, methods for handling noisy or redundant cases, and techniques for maintaining both competence and efficiency as the case base evolves.

\textbf{LLM-Enhanced CBR Processes:} Further investigation into using LLMs for various stages of the CBR cycle presents significant opportunities, including more nuanced case indexing, sophisticated similarity assessment beyond embedding comparisons, and advanced adaptation techniques leveraging the generative capabilities of these models.

\textbf{Multi-Agent CBR Architectures:} Exploring distributed CBR approaches across multiple specialized LLM agents could enable more scalable and robust problem-solving. Research on communication protocols and knowledge sharing between case-based agents could lead to emergent problem-solving capabilities beyond those of individual agents \citep{wu2023autogen}.

\textbf{Robust Evaluation Frameworks:} Developing evaluation frameworks and metrics specifically designed for LLM agents utilizing CBR is essential. These should consider reasoning depth, explanation quality and transparency, adaptation to novel situations, and goal reasoning capabilities.

\textbf{Ethical Considerations:} Exploring ethical implications of using CBR to enhance LLM agents is paramount, including considering potential biases in the case base, ensuring fairness in decisions, and maintaining transparency about how past experiences influence agent behavior.

\section{Conclusion}
\label{sec:conclusion}

This paper introduced a theoretical framework for integrating Case-Based Reasoning (CBR) into Large Language Model (LLM) agents, aiming to improve their reasoning abilities, adaptability, and transparency. By combining CBR with neural language models, we showed how these systems can benefit from both past experiences and learned language patterns. We explored how incorporating cognitive mechanisms like self-reflection, introspection, and curiosity, driven by goal-oriented autonomy, can further deepen an agent's reasoning and knowledge understanding. Our comparison with Chain-of-Thought reasoning and standard Retrieval-Augmented Generation showed that CBR-augmented agents offer clear advantages in reasoning transparency, domain adaptation, and solution quality, particularly in specialized domains requiring structured procedural knowledge. The explicit organization of past experiences in case libraries enables faster learning and provides precedent-based explanations, which are vital for building user trust and improving system interpretability. The theoretical formulations and proposed architectural frameworks presented in this work provide a clear foundation for the future development and implementation of CBR-augmented LLM agents across a wide range of applications. By effectively merging symbolic and neural AI paradigms, this research contributes to the ongoing evolution of hybrid AI architectures that leverage the unique strengths of different approaches. As language models continue to improve, combining them with structured reasoning methods like CBR offers a promising pathway towards more robust, transparent, and adaptive agent systems. Future research into areas like meta-case learning, applying knowledge across different fields, using diverse data types for cases, and improving goal reasoning will push these hybrid systems even further, advancing artificial intelligence and how humans and AI work together.

\bibliographystyle{unsrtnat}
\bibliography{references} 

\end{document}